\documentclass[letterpaper, 10 pt, conference]{ieeeconf}  % Comment this line out
\IEEEoverridecommandlockouts                              % This command is only
\usepackage{graphicx} % for pdf, bitmapped graphics files
\usepackage{amsmath} % assumes amsmath package installed
\usepackage{amssymb}  % assumes amsmath package installed
\usepackage[boxruled,vlined,linesnumbered]{algorithm2e}  % algs
\usepackage{courier}   % for courier text
\usepackage{multirow}  % for tables
\usepackage{hyperref}  % for URLs
\usepackage{booktabs}  % for tables
\usepackage{tabularx}
\usepackage{flushend}  % flushing
\usepackage{bbm}  % for indicator

\SetKwInOut{Parameter}{Parameters}
\SetKwInOut{Input}{Inputs}
\SetKwInOut{Output}{Outputs}

\title{\LARGE \bf
Introducing DeepBalance: Random Deep Belief Network Ensembles to Address Class Imbalance
}

\author{Peter Xenopoulos$^{1}$ \\
{\small peter.xenopoulos@pomona.edu}% <-this % stops a space
\thanks{$^{1}$P. Xenopoulos is a student in the mathematics and economics departments at
        Pomona College in Claremont, CA 91711}%
}

\begin{document}

\maketitle
\thispagestyle{empty}
\pagestyle{empty}

%%%%%%%%%%%%%%%%%%%%%%%%%%%%%%%%%%%%%%%%%%%%%%%%%%%%%%%%%%%%%%%%%%%%%%%%%%%%%%%%
\begin{abstract}

Class imbalance problems manifest in domains such as financial fraud detection or network intrusion analysis, where the prevalence of one class is much higher than another. Typically, practitioners are more interested in predicting the minority class than the majority class, as the minority class may carry a higher misclassification cost. However, classifier performance deteriorates in the face of class imbalance as oftentimes classifiers may predict every point as the majority class. Methods for dealing with class imbalance include cost-sensitive learning or resampling techniques. In this paper, we introduce DeepBalance, an ensemble of deep belief networks trained with balanced bootstraps and random feature selection. We demonstrate that our proposed method outperforms baseline resampling methods such as SMOTE and under- and over-sampling in metrics such as AUC and sensitivity when applied to highly imbalanced financial transaction data sets. Additionally, we explore performance and training time implications of various model parameters. Furthermore, we show that our model is easily parallelizable, which can reduce training times. Finally, we present an implementation of DeepBalance in \textbf{\textsf{R}}.

\end{abstract}

%%%%%%%%%%%%%%%%%%%%%%%%%%%%%%%%%%%%%%%%%%%%%%%%%%%%%%%%%%%%%%%%%%%%%%%%%%%%%%%%
\section{Introduction}

When solving practical classification problems, a practitioner may be faced with class imbalance, meaning that one class has a significantly higher prevalence than the others (also called the \textit{majority} class). Examples of imbalanced classification problems in the literature include \cite{kubat1998machine,kubat1997learning,burez2009handling,wang2013using}. Class imbalance problems may be exacerbated in the future as we discover new methods to collect rare data and rate of data collection increases. In many class imbalance problems, the minority class is not only the interest, but also carries the higher misclassification cost, which complicates learning \cite{chawla2009data}.

Machine learning classifiers try to find an optimal decision boundary that fits training data. As classifiers generally seek to find the simplest rule that partitions the training data, the simplest rule in imbalanced settings is often always predicting the majority class \cite{akbani2004applying}. Results can be deceptive for such classifiers, as they may achieve high accuracy. For example, in a problem where a minority class occurs 0.1\% of the time, an uninformed classifier can achieve 99.9\% accuracy by simply always predicting observations as the majority. Thus, the naturally occurring target class distribution is not optimal for learning in highly imbalanced scenarios \cite{weiss2001effect,japkowicz2000class,japkowicz2002class,he2009learning}. 

Two common methods exist for dealing with class imbalance: resampling methods that balance the data before model training or cost-sensitive learning methods that adjust the relative costs of the errors during model training. While cost-sensitive methods have been shown to alleviate learning difficulties for neural networks in the wake of class imbalance, we investigate the use of resampling methods, as cost-sensitive methods may be hard to implement for practitioners or error costs may be unknown \cite{zhou2006training}. Examples of resampling methods include under- and over-sampling, SMOTE and ROSE \cite{drummond2003c4,chawla2002smote,lunardon2013package}. While under- and over-sampling are some of the simplest methods to implement, they also carry some drawbacks. Under-sampling may discard useful information in an effort to create a balanced data set, and over-sampling may increase the likelihood of model overfitting \cite{mollineda2007class,batista2004study,kotsiantis2006handling}. SMOTE builds upon these two methods by up-sampling the minority class and down-sampling the majority class. Further research related to SMOTE, such as SMOTEboost and Borderline-SMOTE, have shown its ability to increase the performance of underlying classifiers when applied to imbalanced class problems \cite{han2005borderline,chawla2003smoteboost}. These sampling techniques have also been shown to increase the performance of support vector machines across a variety of problems by reducing bias towards the majority class \cite{batuwita2013class,tang2009svms}. However, beyond SVMs and tree-based methods, the effects of class imbalance are still relatively unexplored \cite{guo2008class,van2007experimental}. 

In this paper, we focus on prediction improvements through resampling methods by applying ensemble methodology similar to balanced random forests or \texttt{EasyEnsemble} \cite{chen2004using,liu2009exploratory}. Part of the power of these methods revolve around ensemble learning \cite{sollich1996learning}. Specifically, the constituent base classifier in our ensemble is the deep belief network (DBN). Previous studies have shown that simple approaches which combine random undersampling with bagging or boosting ensembles are favorable in the wake of class imbalance \cite{galar2012review}. Furthermore, these studies noted a particular synergy between random undersampling and bagging. Here, we present DeepBalance, an ensemble of deep belief networks, trained with random feature selection on balanced boostraps. Additionally, we provide an implementation of DeepBalance in \textbf{\textsf{R}}. Finally, we show that DeepBalance is easily parallelizable, making training times more palatable to practitioners. 

The rest of the paper is as follows. In section II, we present related mathematical background and our algorithm, DeepBalance. In section III we outline our experimental methods. In section IV we present our results. In section V we present a discussion of our results and in section VI we conclude the paper.
        
%%%%% BALANCED ENSEMBLE %%%%%
\section{Methods}
\subsection{Class Imbalance}
Suppose we have a classifier, $C$, which separates a space into two. Let us also assume that $C$ is designed to minimize error rate, and thus maximize accuracy, defined as 

\begin{equation} \label{eq:1}
\frac{Misclassifications}{Observations}
\end{equation}

In an imbalanced case, the loss may be dominated by misclassification of the majority class.

\begin{figure}
\centering
\includegraphics[height=3.0in,width=3.0in]{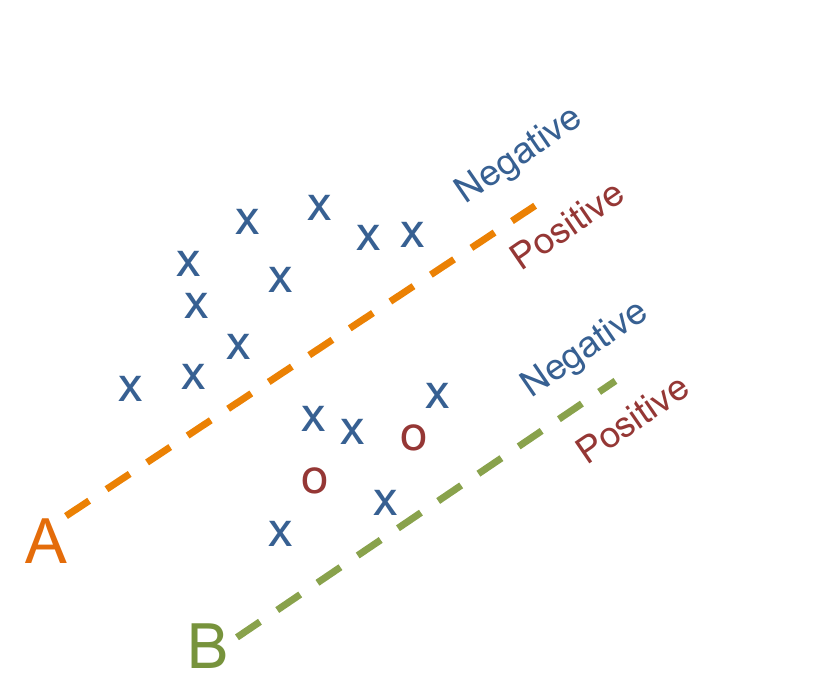}
\caption{An illustrated class imbalance problem.}
\end{figure}

In Figure 1, we have two classifiers, $A$ and $B$. We see that the misclassification error of A is $5/17$, as we misclassify the five blue x's on the "Positive" partition of the classifier. We also see that the misclassification error of B is $2/17$. Thus, we would choose classifier B to maximize our accuracy, even though we misclassify all of the positive cases (red points). 

As mentioned previously, there are two approaches to deal with this imbalance: cost-sensitive learning and resampling methods. Although seemingly different, these two approaches are quite similar in that both seek to adjust for the different costs associated with errors. Cost-sensitive learning changes the loss function to account for the usually higher cost associated with misclassifying a minority class as the majority. One such modification could include adding a penalty for each positive case missed. 

However, dealing with loss functions can be hard for practitioners, as they may lack the ability to directly modify or determine costs. Resampling indirectly adjusts for costs by evening out the class distributions and requires no modification of the loss function. For example, oversampling of the minority increases the cost of misclassifying a minority observation by sampling the same points multiple times. If you oversample an original observation, you now have two of the same observation in your training set. Thus, if your classifier misclassifies that point, it misclassifies it twice.

\subsection{Deep Belief Networks}

The base learner of our proposed ensemble is the deep belief network (DBN) architecture described in \cite{hinton2006fast,hinton2006reducing}. A DBN is composed of multiple restricted Boltzmann machines (RBMs) shown in Figure 2. These RBMs are stacked on top of each other, taking their inputs from the hidden layer of the previous RBM. We use DBNs as our base learners due to their ability to model complex interactions and strong performance at classification tasks \cite{mohamed2009deep,krizhevsky2010convolutional,sarikaya2014application}. Additionally, DBN ensembles have been employed to forecast time series, diagnose failures and to assess credit risk \cite{qiu2014ensemble,zhang2015deep,yu2016novel}.

\begin{figure}
\centering
\includegraphics[height=2.65in,width=3.45in]{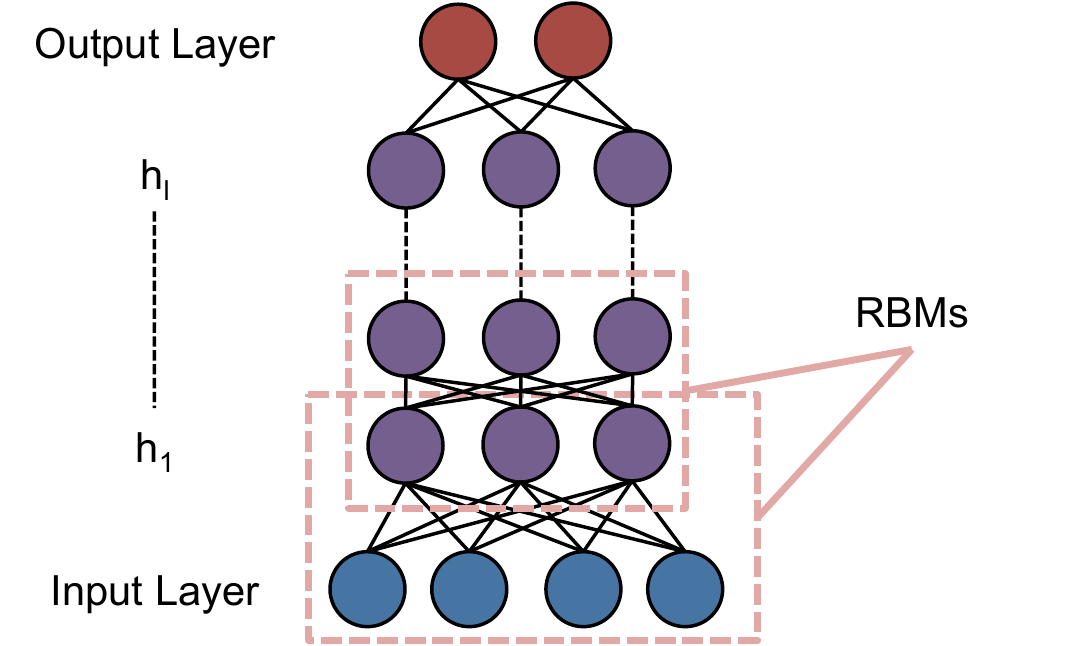}
\caption{An example of a basic deep belief network for binary classification with $k$ hidden layers}
\end{figure}

%%%%% BALANCD ENSEMBLE %%%%%
\subsection{DeepBalance}
We present DeepBalance in Algorithm 1. DeepBalance trains DBNs on balanced bootstraps of data, meaning the bootstrapped data is sampled such that it has equal target class prevalence. The idea of a balanced bootstrap stems from the possibility of bootstrapping samples with no minority cases, which may be especially prevalent in situations with a large class imbalance.

Each DBN is then added to our ensemble. \texttt{TrainDBN} represents the function we use to train a DBN and takes parameters $\varphi$, $\theta$, $mtry$ and $max.it$. $\varphi$ and $\theta$ represent the minority-class and majority-class data, respectively. Each model is trained with at most $mtry$ number of randomly sampled (with replacement) features from the feature space, and the sampled features are denoted as $F_S$. $max.it$ specifies the maximum number of fine-tuning epochs in training a DBN. Each DBN is fine-tuned with backpropagation.

By introducing randomness in the feature selection process, as well as the sampling process, we create a more diverse ensemble, and we can help the ensemble generalize. If one only introduced randomness via the balanced bootstrap process, and the train majority data was very homogenous, the trained classifier may not generalize well to new data.

\begin{algorithm}
\SetAlgoLined
 \Input{Minority-class training set, $\varphi$, Majority-class training set, $\Theta$, where $|\varphi| < |\Theta|$, Feature space, $F$}
 \Output{An ensemble of deep belief networks}
  \Parameter{$mtry$, the number of randomly selected features to try for each model, $total.nets$, the total number of models to include in the ensemble, $max.it$, the total number of epochs for training}
 ${\tt TrainDBN}(\varphi, \theta, F_S, max.it)$\\
 \For{$m \in total.nets$}{
 Sample with replacement $\theta$ from $\Theta$, where $|\theta| = |\varphi|$\\
 Sample $F_S$ from $F$, where $|F_S| = mtry$ \\
 ${\tt TrainDBN}(\varphi, \theta, F_S, max.it)$\\
 Add trained DBN to list of models
 }
 \caption{DeepBalance}
\end{algorithm}

\section{Methods}
\subsection{Data}
To benchmark DeepBalance we use financial transaction data, which is typically imbalanced. While most transactions are not fraudulent, there exist a small percentage that are fraudulent. These few fraudulent transactions usually incur costly monetary repercussions. Thus, it is important for financial transactions to be quickly and accurately classified. Our selected data necessitate the use of resampling methods as conventional train/test splits may not capture enough minority cases to train any effective classifier.

Our first data set contains hundreds of thousands of European credit card transactions made over two days in September 2013 \cite{dal2015calibrating}. It contains 29 features: $V_1, V_2, \ldots, V_{29}$, the principal components of a prior PCA transformation, and $Amount$, the dollar amount of the transaction. We drop the $Time$ variable from the original data set as it represents time since last transaction and mostly acts as an index for the data. In total we have $n = 284,807$ observations, with 492 positive fraud cases, for a fraud incidence of 0.173\%. 

We also chose to benchmark DeepBalance on a simulated financial transaction data from the PaySim simulator \cite{lopez2016paysim}. PaySim is an agent-based simulation which closely mirrors financial transaction data. From this simulated data, we have $n = 6,362,620$ simulated financial transactions where only $8,213$ are fraudulent, for a fraud incidence of 0.129\%. We use the five variables described in Table I to classify the transactions.

% \multicolumn{1}{m{3cm}|}{This is justified and may go to second line as well, neatly}
\begin{table}[]
\centering
\caption{PaySim Features}
\label{paysim-data}
\begin{tabular}{@{}ll@{}}
\toprule
Features       & Description                                                                 \\ \midrule
type           & \multicolumn{1}{m{2.45in}}{Type of transaction, such as CASH-IN, CASH-OUT, DEBIT, PAYMENT and TRANSFER} \\
amount         & Amount of transaction                                                       \\
oldbalanceOrig & Initial balance before the transaction from origin                        \\
newbalanceOrig & New balance after the transaction from origin                               \\
oldbalanceDest & Initial balance recipient before the transaction.                           \\
newbalanceDest & New balance recipient after the transaction.                                \\ \bottomrule
\end{tabular}%
\end{table}

\subsection{Sampling Methods}
We compare various sampling methods against DeepBalance, which we describe in Table II. We institute a 70\%/30\% train/test split on each data set. Specifically, we use 70\% of our positive and 70\% of our negative cases, respectively, to create a training data set. The remaining 30\% of each case is combined to create a testing data set. Our DeepBalance method is parameterized with $total.nets = 25$ and $max.it = 50$. For the European credit card fraud data, we set $mtry = 5$, and for the PaySim data, we set $mtry = 3$. For under-sampling, over-sampling and SMOTE, we trained a DBN using all available features and set $max.it = 100$.

\begin{table}[]
\centering
\caption{Resampling methods}
\label{resampling-methods}
\begin{tabular}{@{}ll@{}}
\toprule
Method         & Description    \\ \midrule
DeepBalance    & \multicolumn{1}{m{2.15in}}{Uses the resampling methods described in methods}  \\
& \\
Over-sampling  & \multicolumn{1}{m{2.15in}}{Over-samples $x$ positive cases to create a balanced sample with $x$ positive and $x$ randomly sampled with replacement negative cases, where $x = 1000$ for the credit card fraud data and $x = 12,000$ for the PaySim data} \\
& \\
Under-sampling & \multicolumn{1}{m{2.15in}}{Under-samples $|\phi|$ majority cases and combines with all of the minority cases} \\
& \\
SMOTE  & \multicolumn{1}{m{2.15in}}{Uses the methodology in,\cite{chawla2002smote} and the implementation in \cite{dmwr}}  \\
& \\
No resampling  & \multicolumn{1}{m{2.15in}}{Uses no resampling method, uses the training data containing 70\% of each class} \\
& \\
All Features Ensemble &  \multicolumn{1}{m{2.15in}}{Similar to DeepBalance, except without random feature selection} \\ \bottomrule
\end{tabular}
\end{table}

\subsection{Performance Metrics}
For imbalanced class problems, accuracy is an inappropriate tool to gauge model performance. A classifier which always predicts the majority class can perform extremely well on an accuracy basis, if the imbalance is high. Common metrics for assessing classifier performance in binary classification problems include the true negative rate ($Acc^-$) and the true positive rate ($Acc^+$), shown below:

\begin{equation}
\begin{split}
Acc^- & = \frac{TN}{TN+FP} \\
Acc^+ & = \frac{TP}{TP+FN}
\end{split}
\end{equation}

The true positive rate is also often referred to as sensitivity. The true negative rate is often referred to as specificity. Additionally, weighted accuracy adds another dimension to performance evaluation, where weighted accuracy is defined as 

\begin{equation}
\beta \times Acc^- + (1-\beta) \times Acc^+
\end{equation}

In this study, we set $\beta$ equal to 0.5. When $\beta = 0.5$, weighted accuracy is sometimes called balanced accuracy.

Additionally we look at the area under the curve (AUC), which we derive from the receiver operating characteristic curve. The receiver operating characteristic curve describes the ability of a classifier to identify both positive and negative classes as the decision threshold changes \cite{davis2006relationship}. 

\subsection{R Implementation}
We offer an implementation of the aforementioned DeepBalance in \textbf{\textsf{R}} \footnote{Our implementation of DeepBalance is currently available in the following GitHub repository: \url{https://github.com/peterxeno/DeepBalance}.}. The implementation of deep belief networks is provided by \cite{dreespackage}. The main functionality of our \textbf{\textsf{R}} implementation is through the \texttt{RNNE()} function, which takes arguments $formula$, $train$, $mtry$, $total.nets$, $max.it$. The $formula$ parameter contains an R formula of a target variable and its predictors, and is typically structured as $Outcome \sim V_1 + V_2 + ... + V_n$ where $V_n$ represents the $n^{th}$ feature. The $train$ parameter represents a data set that contains both $Outcome$ and $V_1, ..., V_n$. The $mtry$, $total.nets$ and $max.it$ parameters are the same as described in section II. Using the features passed through $formula$, the data is separated into data frames containing the majority and minority data (denoted as $\Theta$ and $\varphi$). An empty list is then created to contain all of the trained models. 

In the model training process, we start by sampling with replacement the number of minority cases from $\Theta$, which we call $\theta$, where $|\theta| = |\varphi|$. We then combine $\theta$ and $\varphi$ to create our training set. Then, we sample $mtry$ variables, with replacement, from our predictors. Finally, we train a deep belief network given our combined data, our randomly sampled features and $max.it$. This network is then added to the list of DBNs we initialized earlier. This process is repeated until the length of the list of DBNs is equal to $total.nets$. Thus, our list of DBNs is our ensemble.

\begin{figure*}
\centering
\includegraphics[width=3in,height=3in]{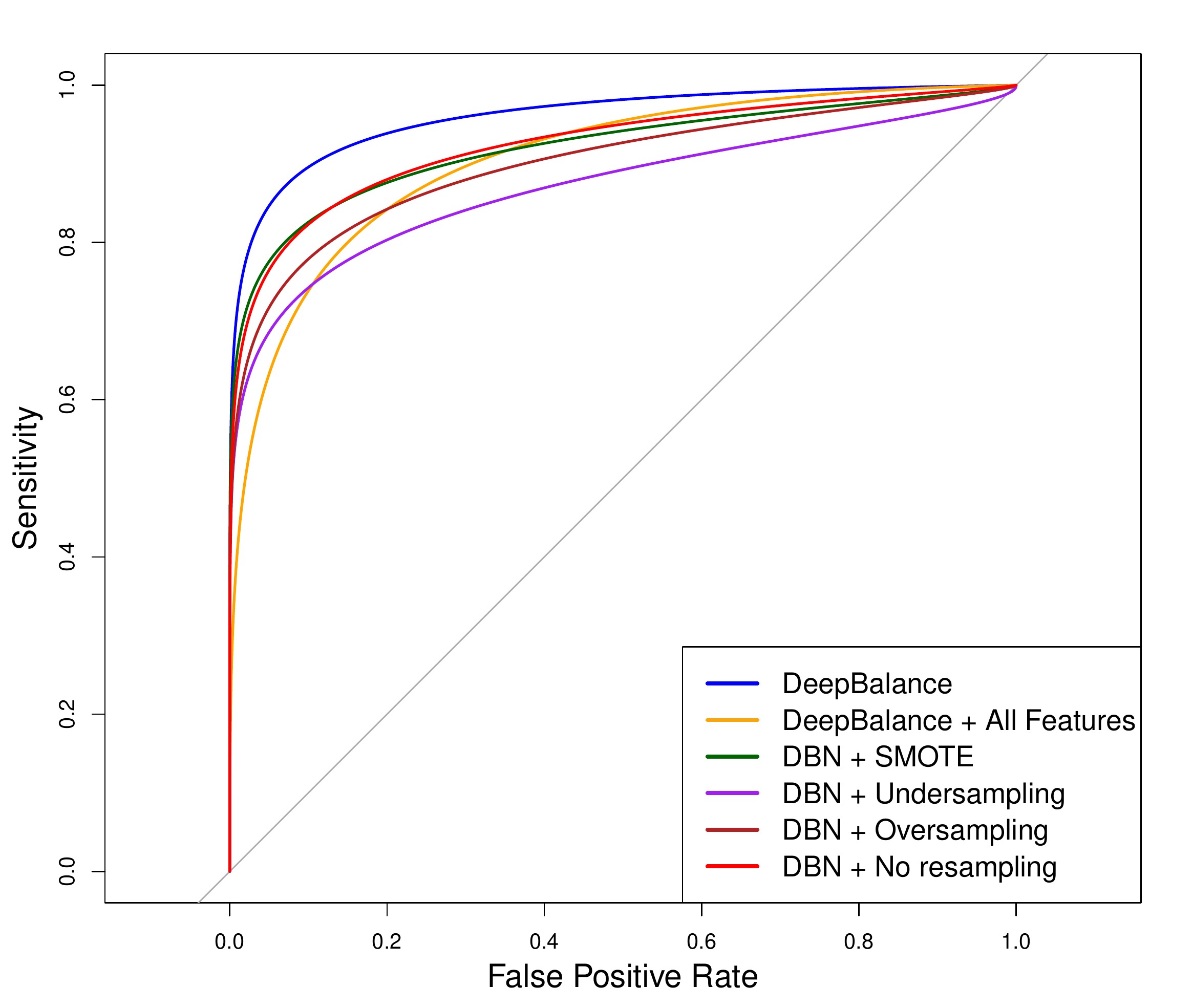}
\includegraphics[width=3in,height=3in]{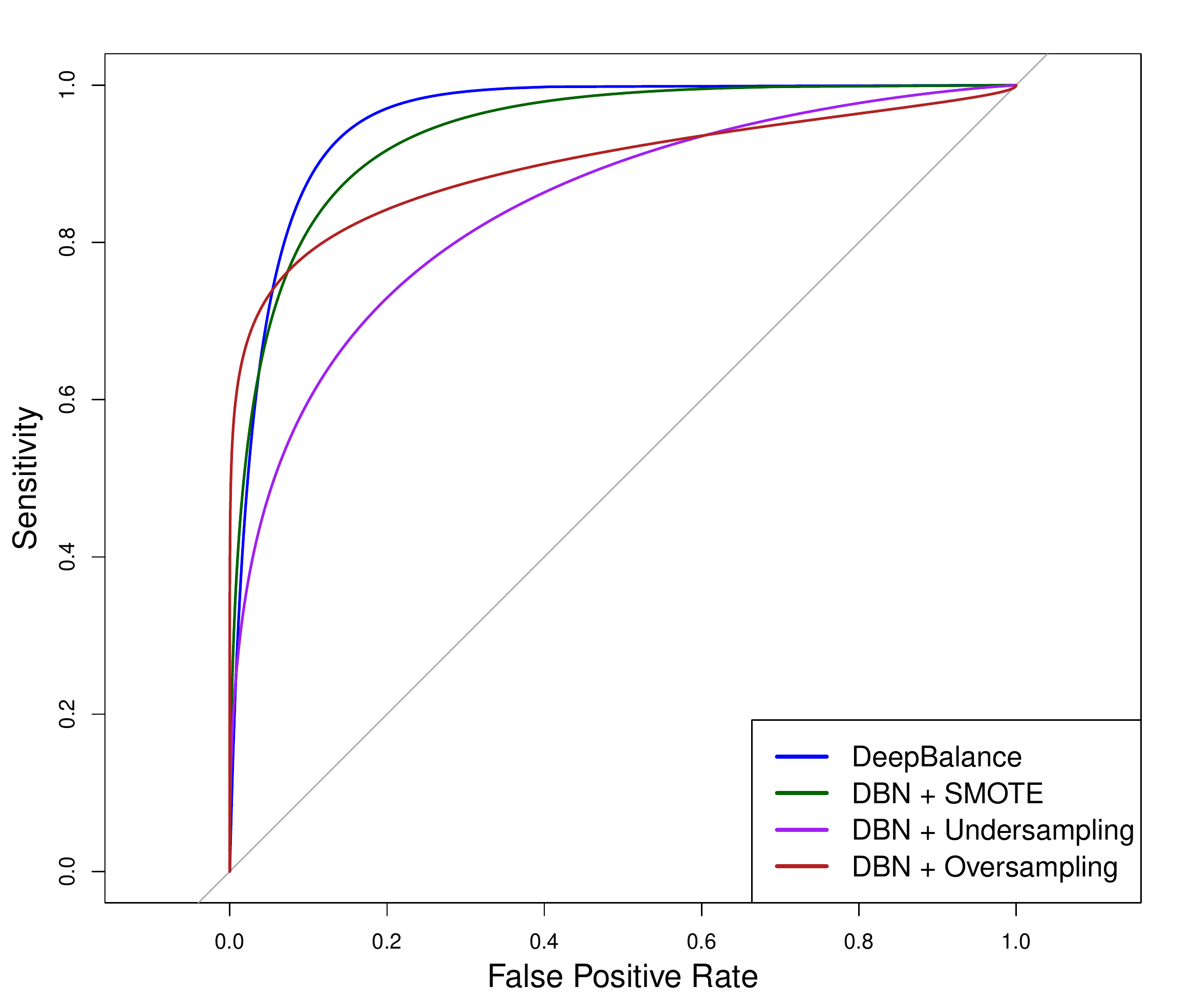}
\caption{Smoothed ROC Curves for European credit card fraud (left) and PaySim (right)}
\end{figure*}

\section{Results}

We present a comparison of DeepBalance versus standard resampling methods for dealing with class imbalance in both Table III and Table IV. We can observe the smoothed ROC curves for both the European credit card fraud and PaySim data in Figure 3. To calculate $Acc^+$, $Acc^-$ and Balanced Accuracy, we assume a decision threshold of 0.5 for both data sets. Bold indicates the best model by respective performance metric. We do not include DBN + No resampling output for PaySim due to restrictive training times (over 4 million observations). We also do not include DBN + All Features for PaySim as its performance converges to the DeepBalance solution.

\begin{table}[]
\centering
\label{results-table}
\caption{Model comparisons (European credit card fraud)}
\begin{tabular}{@{}ccccccc@{}}
\toprule
Method & $Acc^+$ & $Acc^-$ & Balanced Accuracy & AUC \\ \midrule
DeepBalance & \textbf{0.8176} & \textbf{0.9946} & \textbf{0.9061} & \textbf{0.9776}  \\
SMOTE & 0.8176 & 0.8870 & 0.8523 & 0.9158  \\
Undersampling & 0.5338 & 0.9752 & 0.7545 & 0.8248 \\
Oversampling & 0.7500 & 0.9728 & 0.8614 & 0.8672  \\ 
No Resampling & 0.7027 & 0.9707 & 0.8367 & 0.8526  \\ \bottomrule
\end{tabular}
\end{table}

\begin{table}[]
\centering
\label{results-table}
\caption{Model comparisons (PaySim)}
\begin{tabular}{@{}ccccccc@{}}
\toprule
Method & $Acc^+$ & $Acc^-$ & Balanced Accuracy & AUC \\ \midrule
DeepBalance & 0.8766 & 0.9044 & \textbf{0.8905} & \textbf{0.9614}  \\
SMOTE & 0.7618 & \textbf{0.9402} & 0.7311 & 0.9175  \\
Undersampling & 0.8068 & 0.7001 & 0.7534 & 0.7590 \\
Oversampling & \textbf{0.9131} & 0.8169 & 0.8650 & 0.8513  \\ 
\bottomrule
\end{tabular}
\end{table}

We see that DeepBalance outperforms all candidate methods on both data sets. It is important to note the especially strong performance of DeepBalance when applied to the European credit card fraud data set, which may be aided by the features being uncorrelated by virtue of the principal components. 

DeepBalance's performance with the PaySim data illustrates an example of evaluating models across many metrics. While DeepBalance doesn't attain the highest $Acc^+$ or $Acc^-$, we see that it scores the highest in balanced accuracy and AUC, indicating that DeepBalance achieves the best overall performance. Such conclusions would be lost if we only evaluated DeepBalance on an $Acc^+$ or $Acc^-$ basis.

\section{Discussion}

\begin{figure*}
\centering
\includegraphics[width=2.2in]{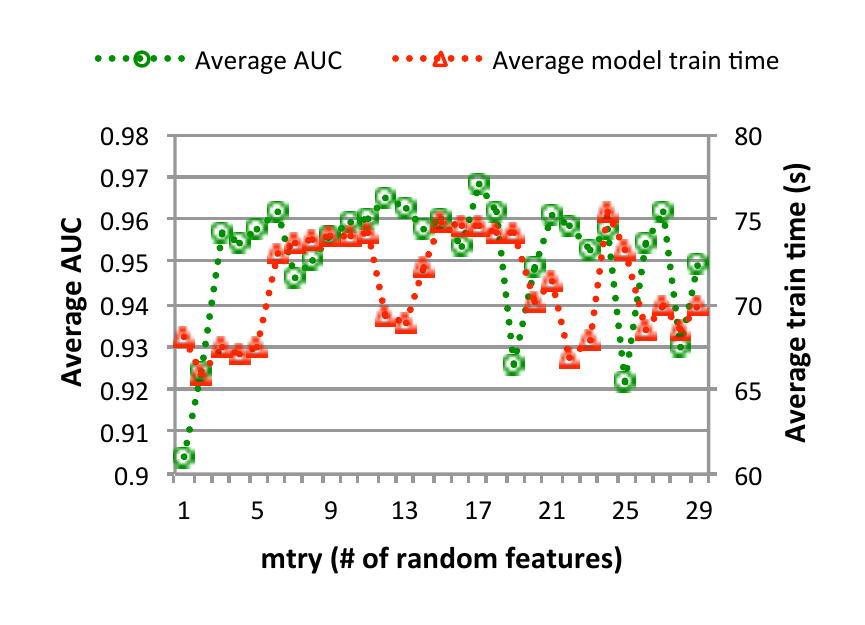}
\includegraphics[width=2.2in]{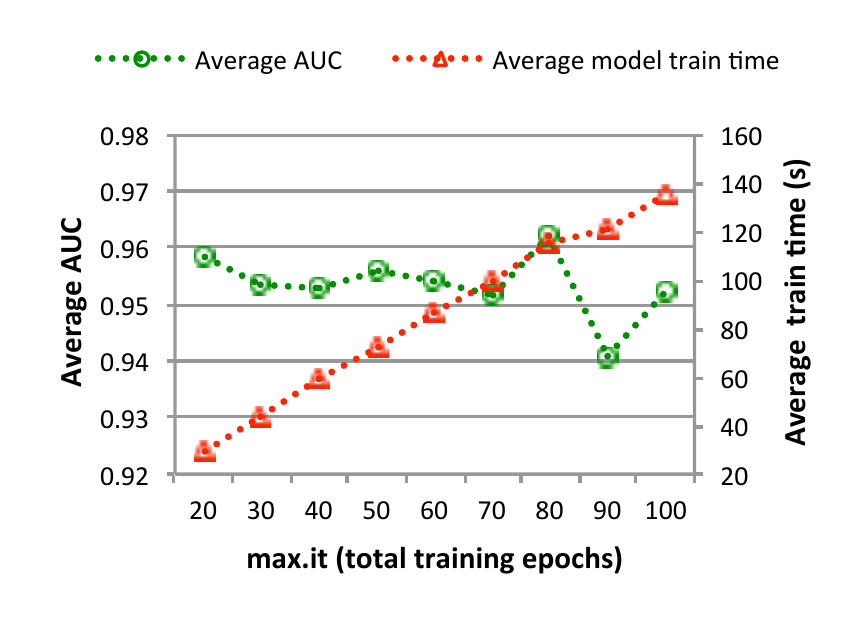}
\includegraphics[width=2.2in]{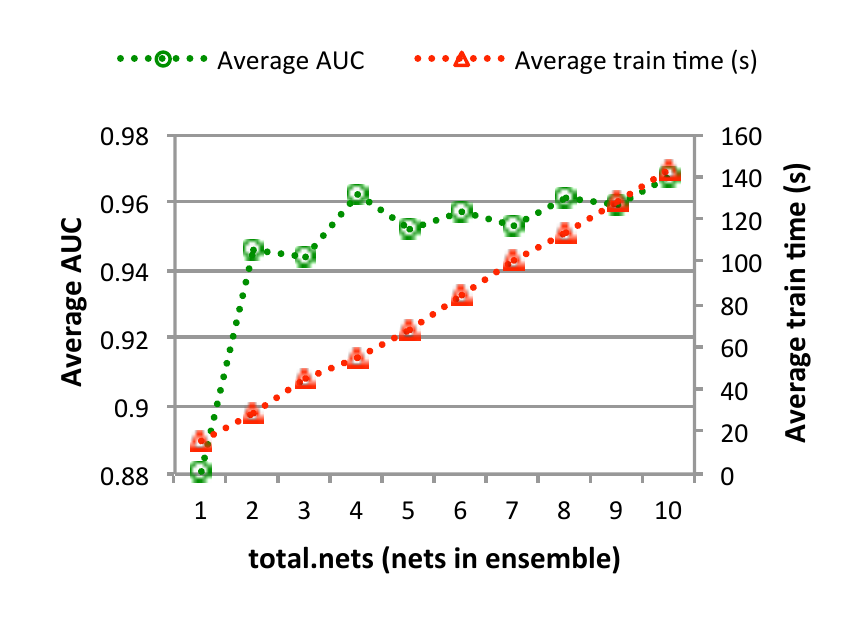}
\caption{Effects of changing $mtry$, $max.it$ and $total.nets$ on model performance (using European credit card fraud data)}
\end{figure*}

While we see that DeepBalance outperforms conventional methods to deal with class imbalance, we are still left uncertain about the properties of its parameters, $mtry$, $total.nets$ and $max.it$. Each of these parameters has implications on model performance and training times. Using the European credit card data, we show how DeepBalance scales in both performance and training time as we change parameters.

We see the effect of increasing $mtry$ on average model performance (measured through AUC) and average model training time in Figure 4. For this experiment, we trained DeepBalance models with $max.it = 50$ and $total.nets = 5$ across $mtry = 1, \dots, 29$ and averaged the model performance at each $mtry$. We used the European credit card fraud data for this experiment. It is interesting to note that there appears to be diminishing returns as $mtry$ grows large. One potential explanation may be that the overall ensemble of models becomes less diverse as each model is using similar sets of features to predict. Thus, keeping a reasonable degree of randomness in feature selection may carry important in performance implications. 

We can observe the effect of increasing $max.it$ on average model performance and average model training time in Figure 4. For this experiment, we trained 5 DeepBalance models with $mtry = 5$ and $total.nets = 5$ across $max.it = 20, 30, \dots, 100$. We then averaged model performance and train times. Overall, we see no gains in model performance, although we see a linear increase in train time. Thus, it may be beneficial for users to set $max.it$ low. Early stopping with backpropagation has been shown to help neural networks generalize better \cite{caruana2001overfitting}.

We can assess the effect of increasing $total.nets$ on average model performance and average model training time in Figure 5. For this experiment, we trained 5 DeepBalance models, each with $mtry = 5$ and $max.it = 50$ across $total.nets$ of $1, 2, \dots, 10$. We see that in general, model performance increases as we add more DBNs to the ensemble. Such model improvement also comes at a training time hit, as it seems that complexity is linear in terms of $total.nets$. However, because training individual nets is an independent process, we can reduce the computational complexity coming from $total.nets$ by introducing parallelism into the algorithm.

While computational complexity may increase through increasing $mtry$, $total.nets$ or $max.it$, we can reduce the model training time through a parallel implementation of DeepBalance. Ensembling lends itself well to parallel implementation because each model in the ensemble is independent of the others. For example, each individual DBN can be trained on a separate processor. The trained models can then be sent back to the master process. If we keep the number of processors, $p$, equal to $total.nets$, then our complexity becomes solely dependent on $mtry$ and $max.it$ as we increase the number of networks we add to our ensemble. Future work will be directed at developing a parallel implementation of DeepBalance.

\section{Conclusion}
In this paper, we presented DeepBalance, an ensemble of deep belief networks trained with balanced bootstraps and random feature selection. Previous literature primarily focused on resampling methods for tree-based methods or cost-based methods. We expanded existing literature by exploring the use of ensembles of deep belief networks in tackling imbalanced class problems in the absence of misclassification costs. We found that DeepBalance achieves slightly better performance than standard resampling methods in performance metrics such as sensitivity, balanced accuracy and AUC when applied to highly imbalanced financial transaction datasets. We believe this to result from the diverse ensemble that DeepBalance creates from both randomness in sampling and in feature selection.

Additionally, we explored the parameters of DeepBalance: $mtry$, $max.it$ and $total.nets$. We see that $total.nets$ has significant  model performance implications, and as $total.nets$ increases, model performance generally increases as ensemble diversity increases. We also observe that $max.it$ and $total.nets$ are linear in training times. The training time increase from training multiple models may be reduced by introducing parallelism into the training process.

Future direction will be directed towards exploring DeepBalance in a multiclass setting, benchmarking DeepBalance on various data sets and investigating other base learners. Ensemble methods have been shown to work well with resampling methods that balance out classes, and we will direct further efforts to a tool which allows the practitioner to select their base learner. Furthermore, we seek to investigate other variable selection techniques, by considering correlation among features in the sampling process or observing feature distributions conditional on the outcome class. 

\addtolength{\textheight}{-12cm}  % This command serves to balance the column lengths

%%%%%%%%%%%%%%%%%%%%%%%%%%%%%%%%%%%%%%%%%%%%%%%%%%%%%%%%%%%%%%%%%%%%%%%%%%%%%%%%

%%%%%%%%%%%%%%%%%%%%%%%%%%%%%%%%%%%%%%%%%%%%%%%%%%%%%%%%%%%%%%%%%%%%%%%%%%%%%%%%

%%%%%%%%%%%%%%%%%%%%%%%%%%%%%%%%%%%%%%%%%%%%%%%%%%%%%%%%%%%%%%%%%%%%%%%%%%%%%%%%

\bibliographystyle{ieeetr}
\bibliography{xenopoulos_ref}

\end{document}